\documentclass[acmsmall,screen,nonacm]{acmart}

\AtBeginDocument{%
  }

\usepackage[ruled,vlined]{algorithm2e}

\usepackage{amsmath,amssymb}
\usepackage{enumitem}
\usepackage{subcaption}
\usepackage{multirow}
\usepackage{placeins}
\usepackage{float}

\newcommand{\attackname}{LeakBoost}
\setcopyright{none}

\usepackage{hyperref}

\begin{document}

\title{LeakBoost: Perceptual-Loss-Based Membership Inference Attack}

\author{Amit Kravchik Taub}
\email{amitkra@post.bgu.ac.il}
\affiliation{%
  \institution{Ben-Gurion University}
  \city{Beer Sheba}
  \country{Israel}
}

\author{Fred M. Grabovski}
\affiliation{%
  \institution{Ben-Gurion University}
  \city{Beer Sheba}
  \country{Israel}
}

\author{Guy Amit}
\affiliation{%
  \institution{Ben-Gurion University}
  \city{Beer Sheba}
  \country{Israel}
}

\author{Yisroel Mirsky}
\affiliation{%
  \institution{Ben-Gurion University}
  \city{Beer Sheba}
  \country{Israel}
}

\renewcommand{\shortauthors}{Kravchik Taub et al.}


\begin{abstract}
Membership inference attacks (MIAs) aim to determine whether a sample was part of a model’s training set, posing serious privacy risks for modern machine-learning systems. 
Existing MIAs primarily rely on \emph{static} indicators, such as loss or confidence, and do not fully leverage the dynamic behavior of models when actively probed. 
We propose \textbf{LeakBoost}, a perceptual-loss-based interrogation framework that actively probes a model’s internal representations to expose hidden membership signals. 
Given a candidate input, LeakBoost synthesizes an \emph{interrogation image} by optimizing a perceptual (activation-space) objective, amplifying representational differences between members and non-members. 
This image is then analyzed by an off-the-shelf membership detector, without modifying the detector itself. 
When combined with existing membership inference methods, LeakBoost achieves substantial improvements at low false-positive rates across multiple image classification datasets and diverse neural network architectures. 
In particular, it raises AUC from near-chance levels (0.53–0.62) to \textbf{0.81–0.88}, and increases TPR@1\% FPR by over an order of magnitude compared to strong baseline attacks. 
A detailed sensitivity analysis reveals that deeper layers and short, low-learning-rate optimization produce the strongest leakage, and that improvements concentrate in gradient-based detectors. 
LeakBoost thus offers a modular and computationally efficient way to assess privacy risks in white-box settings, advancing the study of dynamic membership inference.
\end{abstract}





\maketitle

\section{Introduction}

As machine learning systems are increasingly deployed in practice, concerns have grown regarding the privacy of the training data on which they rely. A central threat in this context is the \emph{Membership Inference Attack (MIA)}, in which an adversary seeks to determine whether a given input was included in the training set of a target model.

This risk is amplified for \emph{image data}, where confirming membership in datasets of faces, medical scans, or personal photographs can directly compromise individual privacy. For example, knowing that a patient’s scan appears in a medical dataset could reveal sensitive health details, or confirming the inclusion of an employee’s identification photo may expose workplace-related attributes. \\
Beyond their real-world impact, image classifiers are also widely used as benchmarks for privacy research, making them a critical domain for evaluating membership leakage and stress-testing defenses.


Prior work distinguishes between black-box and white-box threat models based on the adversary’s level of access to a trained model. 
Among those threat models, the \emph{white-box} setting is especially relevant as it reflects both a powerful adversary and a practical privacy assessment tool. Full access to a model’s internal mechanisms defines an upper bound on achievable membership leakage and provides deeper insight into how memorization manifests within trained models, making this setting particularly well-suited for stress-testing privacy beyond black-box scenarios. Moreover, white-box analysis captures real-world situations in which organizations train models using data that is not fully under their ownership—such as third-party datasets or user-generated content—and seek to ensure that the trained model does not inadvertently disclose the inclusion of specific records. In such cases, understanding the extent to which internal model signals leak membership information becomes crucial for assessing both compliance and potential misuse risks.

While existing MIAs have achieved strong results in both settings \cite{MIAShokri2017, MIANasr2019comprehensive, carlini2022membership, yeom2018privacy}, most rely on \emph{static} indicators such as per-sample loss, or confidence scores. White-box access has enabled the use of richer internal signals, including gradients, influence measures \cite{cohen2024selfinfluence}, and input activations \cite{cretu2024misalignment}. More recent efforts move toward \emph{active probing}, where the target model is deliberately probed to reveal membership cues through adversarial perturbations \cite{del2022leveraging}, gradient inversion \cite{zhu2019deep}, or generative strategies \cite{hitaj2017GAN}.

Others show that \emph{input transformations} alone can amplify leakage without altering the detector \cite{wen2022canary,chen2022membership}. Yet these methods rarely optimize inputs with respect to the model’s \emph{internal feature space}. Perceptual losses—activation-space distances widely used in image synthesis and style transfer—provide a natural measure of representational similarity, but remain underexplored as a probing objective for membership inference.

In this paper, we propose \textbf{LeakBoost}, a perceptual-loss-based membership inference framework that proceeds in two phases. First, in the \textit{interrogation phase}, we synthesize an \emph{interrogation image} by optimizing random noise to match the internal activations of a candidate input under the target model. This optimization serves as a questioning process that amplifies subtle representational differences between members and non-members—effectively pushing the model to “confess” whether the input was in its training set. Second, in the \textit{detection phase}, the interrogation image is passed to a membership \textit{detector} such as GLiR \cite{leemann2023gaussian}, which applies its standard decision rule but now operates on inputs with stronger membership signals.

This paper makes the following contributions:
(i) We introduce \textbf{LeakBoost}, a white-box membership inference framework that actively interrogates a target model by optimizing a perceptual (activation-space) loss prior to detection;
(ii) We evaluate the framework with multiple detectors and show that GLiR forms the most effective pairing, likely due to its gradient-based design;
(iii) We demonstrate that LeakBoost consistently outperforms established white-box baselines on the CIFAR-10 and CIFAR-100 datasets~\cite{krizhevsky2009learning}, with particularly large gains on ViT-4 and AlexNet architectures;
(iv) We conduct a detailed hyperparameter sensitivity analysis, revealing how learning rate, optimization steps, clipping, and layer depth interact with model architecture to shape membership leakage.

\section{Background}

This section provides the technical background necessary to understand our method, covering the notations used for supervised learning, the formulation of membership inference, and the concepts of adversarial examples and perceptual losses that motivate our interrogation approach.

\subsection{Supervised Learning Notation}

Let $\mathcal{D} = \{(x_i, y_i)\}_{i=1}^n$ denote a labeled dataset of input--label pairs with $x_i \in \mathbb{R}^d$ and $y_i \in \{1, \dots, C\}$. 
A neural classifier $f_\theta : \mathbb{R}^d \rightarrow \mathbb{R}^C$ with parameters $\theta$ is trained to minimize an empirical loss
\begin{equation}
    \mathcal{L}_{\text{train}}(\theta) = \frac{1}{n} \sum_{i=1}^{n} \ell(f_\theta(x_i), y_i),
\end{equation}
typically using stochastic gradient descent (SGD) or its variants. 
The model’s performance is evaluated on held-out validation and test sets, while the training set itself remains private. 
Overfitting or memorization during optimization can cause training samples to produce systematically different gradients or activations than unseen data—an effect exploited by membership inference attacks.

\subsection{Membership Inference Attacks}

A membership inference attack (MIA) aims to determine whether a particular input $x$ was part of the training set $\mathcal{D}_{\text{train}}$ of a target model $T_\theta$. 
Formally, the adversary outputs a decision $m(x) \in \{\text{member}, \text{non-member}\}$, or equivalently a score $s(x) \in [0,1]$ that reflects the likelihood of membership. 
Depending on the level of access, two main threat models are studied:

\paragraph{Black-box access.}  
The adversary can only query $T_\theta$ and observe its outputs (labels or confidence scores).  
Attacks exploit differences in prediction loss or confidence between members and non-members~\cite{MIAShokri2017,yeom2018privacy,carlini2022membership,bertran2023quantilemia}.  
Some works extend this approach by training shadow models or analyzing temporal patterns such as loss trajectories~\cite{jia2022trajectorymia}.  

\paragraph{White-box access.}  
The adversary has full access to $T_{\theta}$, including parameters $\theta$, internal activations $e_{\ell}(x)$ and exact input gradients $\nabla_x T_{\theta}(x)$ \cite{MIANasr2019comprehensive}. White-box attacks typically obtain stronger signals but require high privileges.

\subsection{Perceptual Loss and Activation-Space Similarity}

A \emph{perceptual loss}~\cite{johnson2016perceptual} measures similarity between two images in a deep feature space rather than in raw pixels. 
Let $e_\ell(x)$ denote the activation of layer $\ell$ within a network, and let $\mathcal{L}_{\text{perc}}$ denote a weighted sum of feature-space distances:
\begin{equation}
    \mathcal{L}_{\text{perc}}(x', x) = \sum_{\ell \in \mathcal{L}} \lambda_\ell \, d(e_\ell(x'), e_\ell(x)),
\end{equation}
where $d(\cdot,\cdot)$ is a distance metric (e.g., $\ell_2$) and $\lambda_\ell$ are layer weights.  
Prior studies show that memorized samples often exhibit distinctive activation patterns~\cite{arpit2017closer, Stephenson2021Geometry}.

\subsection{Adversarial Examples and Probing Behavior}

Adversarial examples~\cite{szegedy2013intriguing,goodfellow2014explaining,madry2018towards} illustrate that small, targeted perturbations to an input can drastically alter a model’s output. 
Given a loss $\ell(f_\theta(x), y)$, an adversarial perturbation $\delta$ is typically computed by
\begin{equation}
    \delta^* = \arg\max_{\|\delta\|_p \leq \varepsilon} \ell(f_\theta(x + \delta), y),
\end{equation}
where $\varepsilon$ bounds the perturbation size.  
These optimization-based manipulations reveal how models respond to fine-grained input changes—providing a lens into their internal decision surfaces.  
Recent MIA variants~\cite{del2022leveraging,wen2022canary} have leveraged such perturbations as probing mechanisms to elicit membership differences.  

While adversarial optimization is typically used to induce misclassifications, the underlying principle is more general. 
The same mechanism—iteratively modifying an input to optimize a differentiable objective—can be applied to a wide range of analytical purposes. 
Different objective functions can be chosen to probe different aspects of the model’s behavior: one may study its decision boundaries, its feature hierarchies, or its internal representations. 
For example, instead of maximizing a classification loss, one can minimize a \emph{perceptual loss} that measures similarity in the activation space of a neural network, producing an input that reflects how the model perceives a given sample.  

Our approach builds on this intuition but operates in the model’s feature space rather than the pixel space.

\section{Related Work}
\label{related_work}

Existing research on membership inference attacks can be organized along two complementary dimensions: the level of access available to the adversary and the strategy used to extract membership signals.

Along the access-level dimension, a large portion of prior work focuses on the \textit{black-box} setting. In this setting, early work introduced the shadow-model paradigm~\cite{MIAShokri2017}, while subsequent studies demonstrated that simple statistics such as confidence scores or per-sample loss are often sufficient to infer membership~\cite{yeom2018privacy}. More recent black-box attacks adopt statistically grounded formulations, most notably LiRA~\cite{carlini2022membership}, which reframes membership inference as a likelihood-ratio test and achieves strong performance at extremely low false-positive rates, as well as quantile-based approaches that eliminate the need for shadow models by directly modeling non-member score distributions~\cite{bertran2023quantilemia}. 

In the \emph{white-box} setting, where the adversary has access to the model’s parameters, activations, and gradients~\cite{MIANasr2019comprehensive}, richer and more fine-grained membership signals can be exploited. Self-Influence Functions (SIF)~\cite{cohen2024selfinfluence} estimate how much a training example influences its own loss by approximating second-order optimization dynamics, providing a principled but computationally expensive membership score. Cretu et al.~\cite{cretu2024misalignment} introduce \emph{input activations (IA)} as a membership signal, representing the contribution of internal activations to the ground-truth logit and using these features within a shadow-model–based meta-classifier to improve detection. The Gradient Likelihood Ratio (GLiR) attack~\cite{leemann2023gaussian} derives a near-optimal likelihood-ratio test over gradient statistics under Gaussian assumptions, yielding an efficient and theoretically grounded white-box MIA that performs particularly well in the low false-positive rate regime. Finally, LAEQ~\cite{del2022leveraging} exploits differences in adversarial robustness, observing that training members typically require stronger perturbations to cause misclassifications, though this signal is sensitive to regularization and adversarial training. Together, these methods demonstrate the breadth of internal signals that can be leveraged under white-box access, while largely operating on static responses to a fixed input.

While access level determines what information is available, an equally important distinction lies in the strategy used to actively elicit membership leakage.

Beyond static analysis, several works explore \emph{active probing} strategies that manipulate inputs to extract stronger membership cues. Model inversion attacks optimize inputs to maximize class confidence, reconstructing class-level prototypes rather than individual membership~\cite{fredrikson2015inversion}. Gradient inversion methods such as DLG recover training samples by matching gradients through optimization~\cite{zhu2019deep}, while generative approaches train adversarial generators to mimic victim data distributions~\cite{hitaj2017GAN}. Other probing-based attacks leverage training dynamics or adversarial perturbations to expose membership signals~\cite{jia2022trajectorymia,del2022leveraging}. While effective, these approaches are typically tied to specific signals (e.g., gradients or robustness) and do not directly target representational similarity within the network.

A complementary line of work shows that \emph{input transformations} alone can amplify membership leakage without modifying the downstream detector. The \emph{Canary in a Coalmine} attack~\cite{wen2022canary} generates ensembles of adversarially optimized queries around a candidate input and aggregates model responses to improve detection. Chen et al.~\cite{chen2022membership} study scenarios where attacker inputs differ from training samples due to transformations and propose alignment techniques to recover baseline attack performance. However, these methods primarily operate in output space and focus on repairing degraded inputs rather than deliberately amplifying membership-specific differences encoded in the model.

The intersection of these ideas suggests a new avenue for membership inference.  
Static white-box signals reflect a snapshot of model behavior, whereas optimization-based probes reveal how the model \emph{responds} to a candidate input.  
By leveraging perceptual loss as a probing objective, it becomes possible to generate \emph{interrogation images} that align with a model’s internal representations and amplify subtle membership cues.  
This insight forms the basis of our proposed \textbf{LeakBoost} framework.

\section{Threat Model}

This paper considers a \emph{white-box} membership inference setting in which the attacker has full access to a trained target model $T_{\theta}$. The attacker is given a candidate input $x$ and aims to determine whether $x$ was included in the model’s training set $D_{\text{train}}$. We assume that the attacker has complete knowledge of the model architecture and parameters $\theta$, and can perform arbitrary forward and backward passes. In particular, the attacker can extract intermediate activations $e_{\ell}(x)$ from any layer, compute gradients with respect to the input or internal representations, and evaluate loss and gradient statistics on chosen inputs.

The attacker may submit both naturally occurring samples and synthetically generated queries to the target model, and is assumed to have access to auxiliary data drawn from the same input distribution $\mathbb{D}_X$, in order to approximate non-member behavior. However, the attacker does not observe the actual training set and has no direct access to individual training examples. The attack operates solely through analysis of the trained model and its responses to submitted inputs.

This white-box setting captures a strong but realistic adversarial scenario, corresponding to cases such as insider access, debugging or evaluation interfaces, or third-party auditing of models trained on sensitive or partially proprietary data. While powerful, this threat model does not guarantee successful inference in all cases: membership leakage may be reduced by strong generalization, privacy-preserving training mechanisms, or distributional mismatch between candidate inputs and the training data.

\section{LeakBoost Framework}

\subsection{Method Overview}

Neural networks often memorize fine-grained features of training data, but these signals are subtle and hard to detect directly. Our motive is to make the target model \emph{expose} whether candidate inputs are members of its training set, by amplifying those features. To amplify them, we interrogate the model by synthesizing \emph{interrogation images} that align with its internal representations of a query sample.

\begin{figure}[h]
    \centering
    \makebox[\textwidth][c]{%
        \includegraphics[width=1\linewidth]{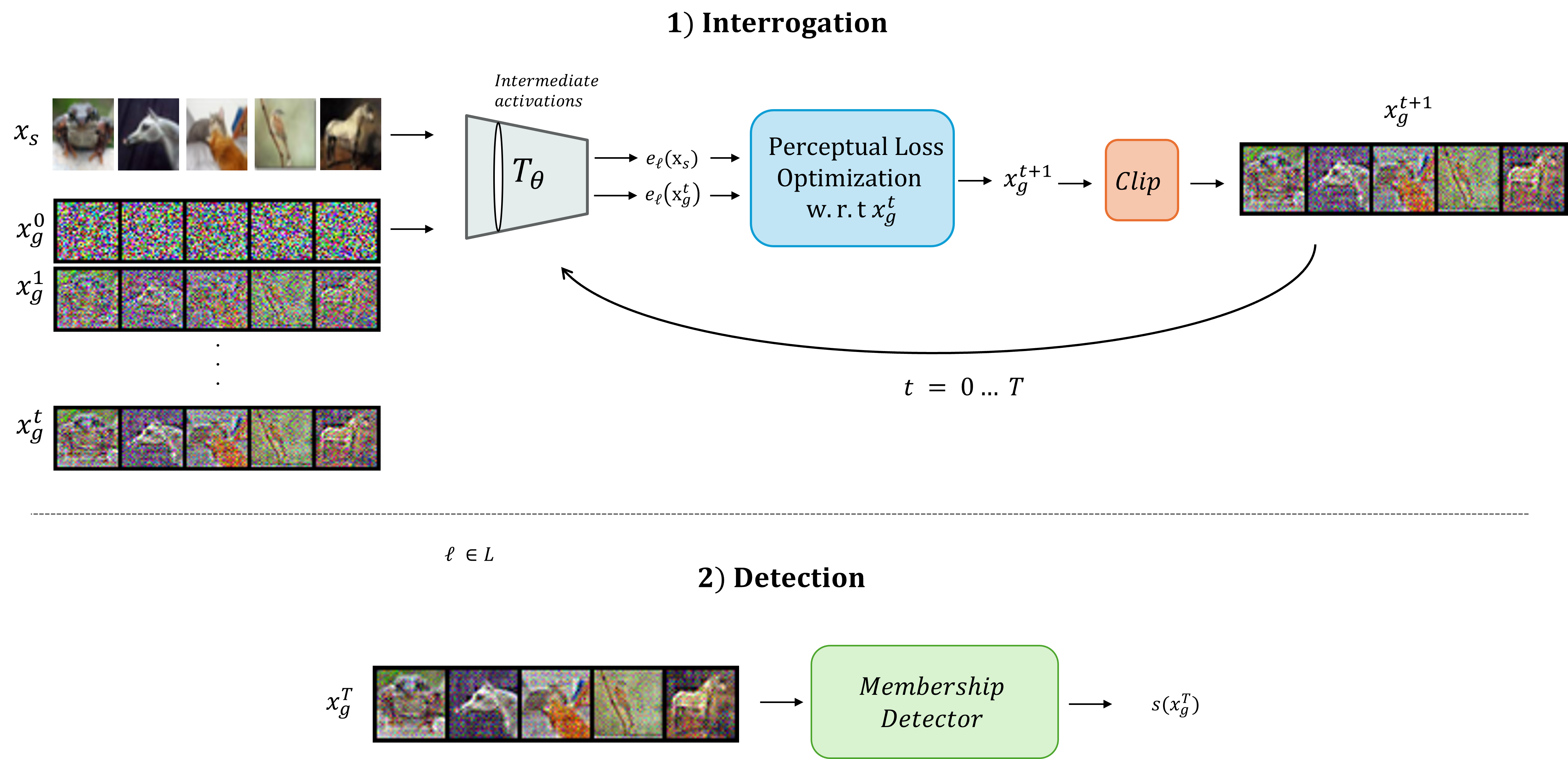}%
    }
    \caption{\textbf{LeakBoost pipeline.} Given a target image $x_s$ and a target model $T_\theta$, optimize $x_g$ for T steps to minimize $\mathcal{L}_{\mathrm{perc}}(x_s;x_g)$ to obtain the interrogation image, which is then passed to a detector to infer membership.}
    \label{fig:leakboost_pipeline}
\end{figure}
These interrogation images act as \emph{boosted samples}:
They amplify differences in the model’s internal representations between training samples and unseen inputs, thereby increasing the separability of members and non-members. We then feed the boosted samples into a membership-inference detector, which can more easily exploit these strengthened signals (see the full pipeline on Fig. \ref{fig:leakboost_pipeline}). In this way, interrogation functions as a modular booster, converting each candidate into a sample explicitly tuned to reveal memorization and expose membership information.

\subsection{Interrogation Image Generation}
\label{subsec:interrogation}
Given a target model $T_{\theta}$ and a query input $x$, we initialize a synthetic image $x_g^{(0)}$ from random noise (e.g., Gaussian). At each iteration $t$, we update $x_g^{(t)}$ by minimizing a perceptual loss that encourages $x_g^{(t)}$ to activate the target model similarly to $x$:
\begin{equation}
\mathcal{L}_{\text{perc}}(x_g, x) = \sum_{\ell \in L} \lambda_{\ell}\, d\big(e_{\ell}(x_g), e_{\ell}(x)\big),
\end{equation}
where $e_{\ell}(\cdot)$ is the internal activation of the model in layer ${\ell}$, $L$ is a chosen set of layers, $\lambda_{\ell}$ are nonnegative weights, and $d(\cdot,\cdot)$ is a distance function such as mean-squared error (MSE). Unlike task losses, this objective does not depend on the class label $y$ but instead probes how the model internally represents $x$.

\begin{figure}[h]
    \centering
    \makebox[\textwidth][c]{%
        \includegraphics[width=1\linewidth]{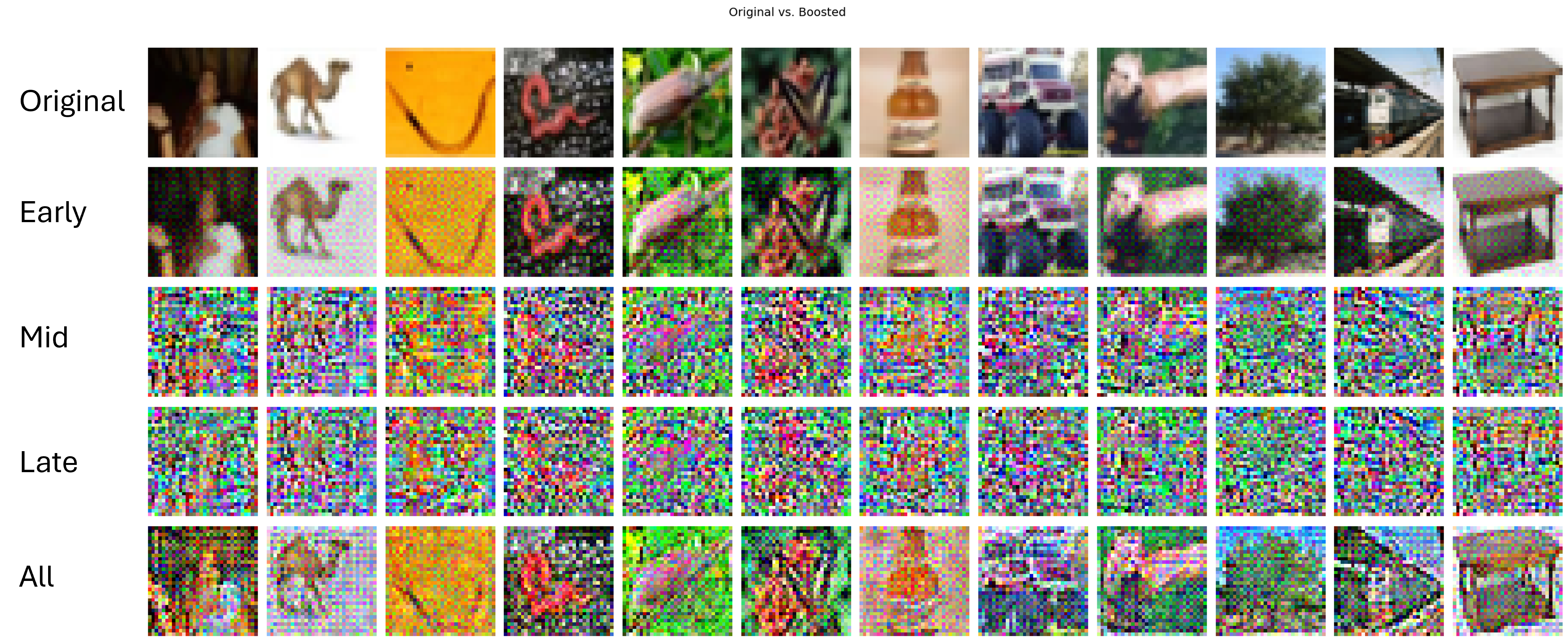}%
    }
    \caption{Examples of interrogation outputs. Each column shows an original image (top) and its boosted counterparts generated by optimizing toward different layer levels of the target model. Boosted images exhibit distinct noise patterns that amplify representational differences between members and non-members.}

    \label{fig:boosting_examples}
\end{figure}

The synthetic image update rule is
\begin{equation}
x_g^{(t+1)} \gets \text{Clip}\Big(x_g^{(t)} - \eta \,\mathcal{O}\big(\nabla_{x_g} \mathcal{L}_{\text{perc}}(x_g^{(t)}, x)\big)\Big),
\end{equation}
where $\eta$ is the learning rate, $\mathcal{O}$ is a first-order optimizer (e.g., SGD, Adam), and \text{Clip} enforces validity constraints on the generated image by projecting it to the empirical dataset distribution (e.g., pixel-wise bounds or normalization consistent with the training distribution). After $T$ steps, we obtain the \textit{interrogation image} $x_g^{(T)}$.
See examples of interrogation outputs in Figure \ref{fig:boosting_examples}, and the full pseudo code in Algorithm \ref{alg:interrogate}.

\begin{algorithm}[t]
\caption{\textsc{Interrogate} — synthesize interrogation image}
\label{alg:interrogate}
\DontPrintSemicolon
\KwIn{
Target model $T_{\theta}$;
query image $x$;
layer set $L$;
layer weights $\{\lambda_\ell\}$;
distance function $d$;
number of steps $T$;
step size $\eta$;
first-order optimizer $\mathcal{O}$;
data-domain clipping operator $\mathrm{Clip}$.
}
\KwOut{Interrogation image $x_g^{(T)}$.}

\BlankLine

\textbf{Initialization.}
Sample $x_g^{(0)} \sim \mathcal{N}(0, I)$ and initialize optimizer state.\;
\textbf{Target activations calculation.}
For each $\ell \in L$, compute $h_{\ell} \leftarrow e_{\ell}(x;\theta)$.\;

\For{$t = 0, \dots, T-1$}{
  \tcp{Perceptual loss}
  $\mathcal{L}_{\text{perc}}^{(t)} \leftarrow
  \sum_{\ell \in L} \lambda_{\ell}\,
  d\!\left(e_{\ell}(x_g^{(t)};\theta),\, h_{\ell}\right)$\;

  \tcp{Gradient-based update}
  $grad^{(t)} \leftarrow \nabla_{x_g} \mathcal{L}_{\text{perc}}^{(t)}$\;
  $x_g^{(t+1)} \leftarrow \mathcal{O}\!\left(x_g^{(t)}, grad^{(t)}, \eta\right)$\;

  \tcp{Data-domain clipping}
  $x_g^{(t+1)} \leftarrow \mathrm{Clip}\!\left(x_g^{(t+1)}\right)$\;
}
\Return{$x_g^{(T)}$}\;
\end{algorithm}

\subsection{Detection}

Once the interrogation image $x_g^{(T)}$ is obtained, we pass it to a \emph{detector} that performs membership inference. We hypothesize that the nature of interrogation makes it particularly effective when paired with detectors that exploit \emph{gradient- and representation-based signals}. 

We focus on the \textbf{Gradient Likelihood Ratio (GLiR)} attack proposed by \citeauthor{leemann2023gaussian}, a theoretically grounded MIA that bypasses the need for shadow models. GLiR frames membership inference as a binary hypothesis test on gradient distributions. 
The following describes the GLiR membership inference attack, which is applied unchanged in our framework.

For a candidate input $x$ with loss function $\ell(\theta;x,y)$ and model parameters $\theta$, consider the gradient vector
\[
g(x) = \nabla_{\theta} \ell(\theta;x,y).
\]
Under the null hypothesis $H_0$ (non-member), $g(x)$ is assumed to follow a Gaussian distribution centered at the population gradient, while under the alternative $H_1$ (member), it is biased towards the empirical training gradient. More formally,
\[
H_0 : g(x) \sim \mathcal{N}(\mu_0, \Sigma), 
\qquad
H_1 : g(x) \sim \mathcal{N}(\mu_1, \Sigma),
\]
where $\mu_0$ and $\mu_1$ denote the expected gradients for non-members and members, respectively, and $\Sigma$ is a shared covariance. GLiR then computes the log-likelihood ratio statistic
\[
\Lambda(x) = \log \frac{p(g(x)\mid H_1)}{p(g(x)\mid H_0)} \;=\; (g(x)-\tfrac{1}{2}(\mu_0+\mu_1))^\top \Sigma^{-1} (\mu_1-\mu_0).
\]
Membership is predicted when $\Lambda(x)$ exceeds a threshold calibrated to the desired false-positive rate.

\subsection{End-to-end attack}
The end-to-end \attackname{} attack (see Alg. \ref{alg:leakboost_e2e}) proceeds in two stages: (i) synthesize an interrogation image $x_g^{(T)}$ for each query $x$ by aligning its internal activations, and (ii) run the detector on $x_g^{(T)}$ exactly as on a natural input. The detector’s method is unchanged—only its input is replaced by a boosted sample that amplifies membership cues—yielding a final score $s(x)$ and decision $\hat m(x)$.

\begin{algorithm}[t]
\caption{End-to-end \attackname{}: Interrogate then Detect (detector unchanged)}
\label{alg:leakboost_e2e}
\DontPrintSemicolon
\KwIn{
target model $T_{\theta}$;
query image $x$;
layer set $L$;
layer weights $\{\lambda_{\ell}\}_{\ell \in L}$;
distance function $d$;
number of steps $T$;
step size $\eta$;
first-order optimizer $\mathcal{O}$;
data-domain clipping operator $\mathrm{Clip}$;
\textsc{Detector} with parameters $\psi$.
}
\KwOut{Membership score $s(x)\in[0,1]$ and decision $\hat m(x)\in\{\text{member}, \text{non-member}\}$.}

\BlankLine
\textbf{1. Interrogation (boosted sample synthesis).}\;
$x_g^{(T)} \leftarrow \textsc{Interrogate}(T_{\theta}, x, L, \{\lambda_\ell\}, d, T, \eta, \mathcal{O}, \mathrm{Clip})$\;

\BlankLine
\textbf{2. Detector on boosted sample.}\;
$s(x) \leftarrow \textsc{Detector}_{\psi}\big(T_{\theta}, x_g^{(T)}\big)$ \tcp*{same detector applied to the original input}
$\hat m(x) \leftarrow \mathbb{I}[\,s(x) \ge \gamma\,]$ \tcp*{threshold chosen for target FPR}

\Return{$(s(x), \hat m(x))$}\;
\end{algorithm}

\section{Experimental Setup}

\subsection{Baselines Implementation}
We evaluate our method against a diverse set of recent \emph{white-box} membership inference attacks. These baselines were chosen because they represent complementary strategies for exploiting internal model information, including gradient-based statistics (GLiR) \cite{leemann2023gaussian}, influence-function analysis (SIF) \cite{cohen2024selfinfluence}, adversarial robustness signals (LAEQ) \cite{del2022leveraging}, and shadow-model meta-classification with input activations (IA) \cite{cretu2024misalignment}. A detailed discussion of these methods and their design is provided in Section~\ref{related_work}.  
We implement each baseline as specified in prior work, and summarize our key implementation choices below. 

\textbf{GLiR.}
We implement GLiR following its formulation as a gradient likelihood-ratio test under Gaussian assumptions. To ensure scalability to high-dimensional models, we subsample gradients to \(d=5000\), apply regularization to the covariance estimate, and compute log-probabilities via a normal approximation to the non-central \(\chi^2\) distribution.

\textbf{SIF.}
We instantiate the self-influence attack so that any misclassified example is automatically treated as a non-member, with augmentation-based averaging disabled, and we use a single recursion depth and one conjugate-gradient iteration to approximate the inverse Hessian--gradient product. Since we evaluate the attack across multiple FPR points, we do not fit fixed thresholds as in the original implementation. Instead, we directly use the SIF scores of member and non-member samples to compute the evaluation metrics.

\textbf{LAEQ.}
We implement the adversarial-distance membership inference strategy, which infers membership by measuring the magnitude of the perturbation required to produce an adversarial example.
While the original authors used the maximum (L$\infty$) adversarial perturbation as the membership inference signal, we found this score to be poorly distributed (often saturating at 0 or 1), which hindered ROC-based evaluation. Instead, we use the L2 norm of the adversarial perturbation, which provides a more continuous and informative signal for computing true positive rates across varying false positive rates.

\textbf{IA.}
For the white-box shadow modeling baseline, we follow the standard threat model where the attacker has access to an auxiliary shadow dataset. We train 5 shadow models in the same way the target model was trained, as described later in this section in \ref{subsec:models}. Since the goal is to train a meta-classifier that learns to distinguish members from non-members based on the outputs of the shadow models, we split the shadows accordingly: one shadow model is reserved to validate the meta-classifier during training, while the remaining shadows provide the meta-training examples. The meta-test set is constructed from the target model’s own member and non-member splits, which are never used in the training phase. 
From each model, we extract the following features:  the activations of the penultimate layer, the loss gradients with respect to these activations, and the ``IA'' (input activation) component, which captures how the activations contribute directly to the prediction of the correct class through the final linear layer. In practice, this provides both the raw representation, the direction in which the loss changes, and the strength of alignment between the representation and the class weights, giving the meta-classifier a rich signal to distinguish members from non-members.
Finally, we train a neural meta-classifier on these features using the Adam optimizer with learning rate $10^{-3}$ (decayed to $10^{-4}$), a batch size of $128$, early-stopping patience of $5$, and a maximum of $30$ epochs.

\subsection{Metrics}

To meaningfully evaluate membership inference attacks, following the convention in recent MIA work \cite{carlini2022membership}, we focus on performance in the \textbf{low–false positive rate regime}, where privacy risks are most critical. 

Therefore, we report \textbf{TPR@low-FPR}, which directly measures how often the attack correctly identifies membership while constraining false alarms to extremely low levels.

In addition, we report \textbf{partial AUC (Area Under the ROC Curve) (pAUC) at low-FPR}, which aggregates the area under the ROC curve restricted to the low FPR region. By summarizing the curve over this range,  we argue that this offers a more stable and informative measure of adversarial advantage in practice, complementing the existing metrics.

For consistency and comparability with prior MIA studies, we also include the \textbf{AUC} as a reference. However, unlike AUC—which reflects performance over the entire ROC curve, including high-FPR regions that are less relevant in privacy settings—we treat it as supplemental. Our primary metrics of interest are TPR@low-FPR and pAUC@low-FPR, which we believe more faithfully reflect adversarial power under realistic privacy constraints.

\subsection{Datasets}
We conduct experiments on the CIFAR-10 and CIFAR-100 datasets \cite{krizhevsky2009learning}, each consisting of 50,000 training and 10,000 test images of size $32 \times 32$. Following the common protocol introduced in \cite{carlini2022membership}, we split the 50,000 training samples into two halves: one half is used to train each model, while the other half serves as the non-member set. All splits are performed in a stratified manner to preserve the original class distribution. Within the training half, $5\%$ of the samples are further reserved for validation, also using stratified sampling. This procedure is repeated independently for each run, so that the training sets of different models are not identical but partially overlapping. We make this assumption primarily due to the small size of the datasets.

\subsection{Models}
\label{subsec:models}
We evaluate our method on four representative image classifiers: ResNet-18 \cite{he2016deep}, AlexNet \cite{krizhevsky2012imagenet}, DenseNet \cite{huang2017densely}, and a lightweight Vision Transformer (ViT-4) \cite{dosovitskiy2020image}. These models capture a broad range of capacities and design paradigms. ResNet-18 serves as a modern convolutional baseline with residual connections widely used in robustness and privacy research. AlexNet, though older and shallower, provides a lower-capacity contrast that remains common in membership inference benchmarks. DenseNet offers a stronger convolutional architecture with densely connected layers, while ViT-4 introduces a transformer-based design that replaces convolutions with self-attention, allowing us to test whether our attack extends beyond CNNs.

Together, these models provide a diverse testbed spanning shallow to deep CNNs and convolutional to transformer-based designs. The models' accuracies are documented in Table \ref{tab:target_models_acc}

Models are trained from scratch using stochastic gradient descent with Nesterov momentum $0.9$ and weight decay $10^{-4}$. The learning rate is set to $0.1$, and the batch size to $100$. We employ a learning-rate schedule that linearly warms up for five epochs and then follows a cosine annealing decay until the end of training. Each model is trained for up to 400 epochs, with early stopping triggered if the validation accuracy does not improve for 20 consecutive epochs. During training, we apply standard data augmentations consisting of random cropping and horizontal flipping, followed by per-channel normalization.
This training setup closely follows the procedure used in LiRA \cite{carlini2022membership}.

\subsection{Interrogation Implementation}
As detailed in Section~\ref{subsec:interrogation}, the interrogation stage synthesizes a boosted input by optimizing a synthetic image to match the internal activations of a given query under the target model.
We next outline the implementation choices used in our experiments.

\begin{description}
    \item[Optimizer.] We use \textbf{Adam} optimizer \cite{kingma2014adam} with fixed betas (0.9, 0.999) and no weight decay, and sweep only the \textit{learning rate}.

    \item[Perceptual loss and layer levels.] We compute the perceptual objective with \textbf{mean squared error (MSE)} over a chosen set of internal layers. For each target model, we evaluate four layer levels:
\emph{Early}, \emph{Mid}, \emph{Late}, and \emph{All}. Each level is a predefined group of layers (equal per-layer weights). The exact layer lists for each architecture appear in \ref{tab:layers_subsets}.

    \item[Initialization and update loop.] For each target sample, we initialize the optimized input 
 $x_g^{(0)}$ as a random Gaussian noise drawn from the valid pixel range and then apply the same normalization used in training. We run \textit{T }optimization steps (swept as a hyperparameter). At every step $t$, we compute the perceptual loss, backpropagate $\nabla_{x_g}$, and update $x_g^{(t)}$ with Adam using the current \textit{learning rate}.
 
     \item[Value clipping vs.\ unconstrained updates.] At the end of each step, we optionally apply \emph{value clipping} to $x_g^{(t)}$ by clamping pixel values to the empirical min/max of the original training distribution for the corresponding dataset. We experiment with both settings:
    clipping is enabled - clamp to training range, and clipping is disabled - allow values to exceed the range. The latter tests whether escaping the training range yields stronger membership signals. No projection onto an $\ell_p$ ball (i.e., no $\varepsilon$-ball constraint) is used in our experiments.

    \item[Attack data splits.] For each target model and dataset, we construct two disjoint splits:
\begin{itemize}[leftmargin=*]
    \item \textit{Attack-validation} (for hyperparameter tuning): \textbf{2K members} and \textbf{2K non-members}. Members are sampled from the model's training set; non-members are sampled from a held-out set disjoint from training. This split is used \emph{only} for hyperparameters tuning and selection.
    \item \textit{Attack-test} (for final reporting): \textbf{10K members} and \textbf{10K non-members} (the original dataset's test set), sampled independently of the validation split. No tuning is performed on this split.
\end{itemize}

\item[Tuning  and evaluation protocol.] All hyperparameter sweeps (values are listed in Table \ref{tab:hyperparameters}) and candidate-signal selection are performed on the \emph{attack-validation} split.
Because our \textit{attack-test} split combines 10,000 member and 10,000 non‐member examples, the lowest false positive rates we can reliably distinguish correspond to\textbf{ 1\% FPR }(100 false positives) and \textbf{0.1\% FPR} (10 false positives). FPR values below 0.1\% become unstable—an extra false positive can dramatically shift TPR, undermining meaningful comparison.
The final configuration is chosen by \textbf{pAUC@1\% FPR} on \textit{attack-validation} and then applied to the \emph{attack-test} split for reporting. 

\end{description}

\begin{table}[h]
\centering
\begin{tabular}{cc}\toprule

\textbf{Hyperparameter} & \textbf{Values} \\\midrule

 Optimization steps& 80,\;120,\;200 \\
Optimizer learning rate & 0.05,\;0.1,\;0.2 \\
Value clipping & True,\;False \\

 Layers&Early,\;Mid,\;Late,\;All\\ 
 Optimizer&Adam\\
 Perceptual loss function&MSE\\\bottomrule
\end{tabular}
\caption{Hyperparameter options explored in our experiments.}
\label{tab:hyperparameters}
\end{table}

\subsection{Experiments}\label{sec:experiments}

\begin{description}
\item[Full attack evaluation (LeakBoost + GLiR).] We evaluate the end-to-end LeakBoost pipeline with GLiR as the detector, comparing it to all baselines (original GLiR, SIF, LAEQ, and IA).

\item[Boosting other detectors and component analysis.] 
We apply the interrogation (boosting) step to alternative detectors (SIF, LAEQ, IA) to measure how much each method benefits from boosted inputs.

\item[Hyperparameter sensitivity.] To examine the robustness of our attack, we perform a hyperparameter sensitivity analysis. 
In this experiment, we vary the main optimization parameters and the layer depth used for interrogation and evaluate their effect on attack performance. 
The analysis is carried out on AlexNet and ViT4, which we found to be the most affected models, 
and provides insight into which design choices most strongly influence the strength of the membership signal.
\end{description}

\section{Experiments Results}

\subsection{Full attack evaluation}
\begin{figure}[h]
    \centering
    \includegraphics[width=1\linewidth]{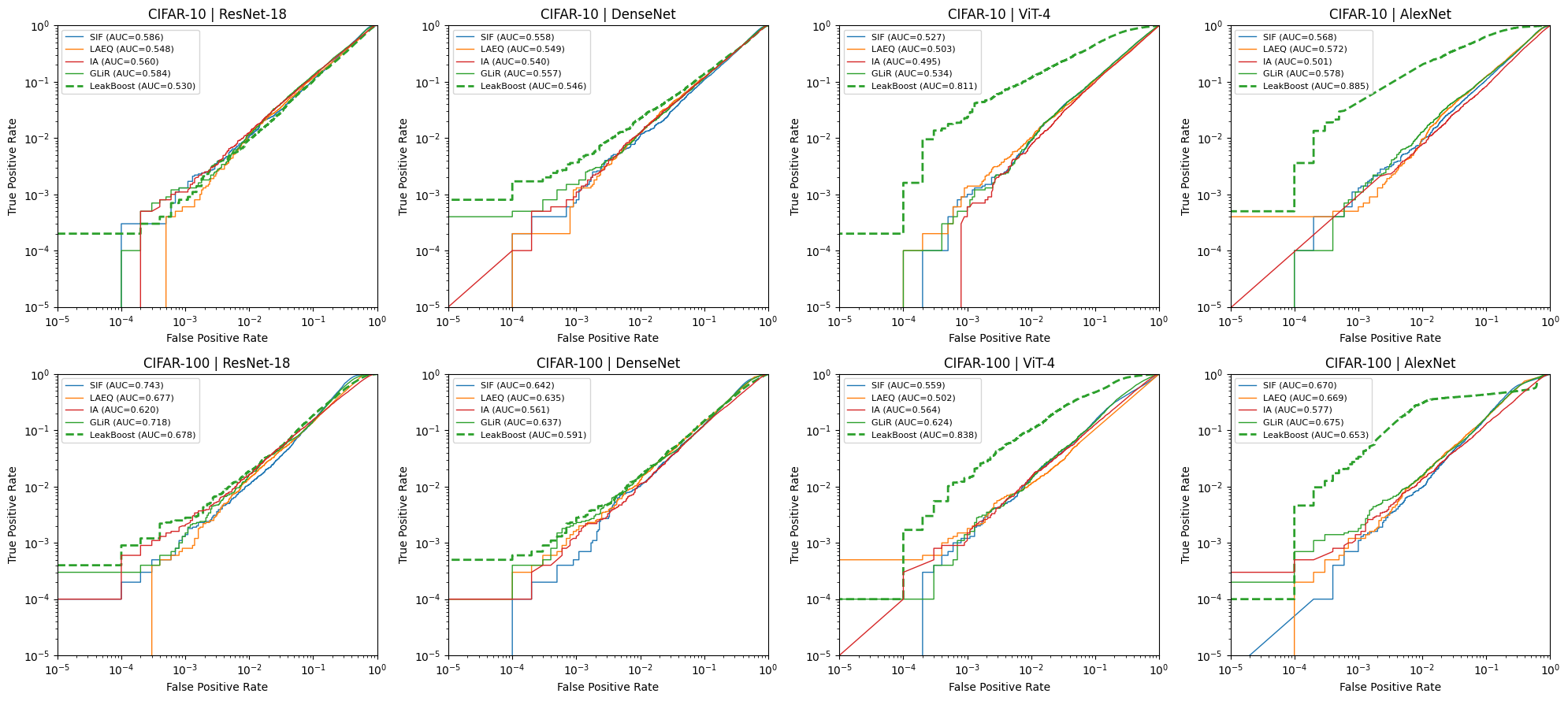}
    \caption{ROC curves for membership inference attacks on CIFAR-10 and CIFAR-100 across four architectures. Solid lines denote baseline methods (SIF, LAEQ, GLiR, IA) and dashed green denotes LeakBoost.}
    \label{fig:leakboost_vs_baselines_rocs}
\end{figure}

We begin by evaluating the end-to-end LeakBoost pipeline with GLiR as the detector, comparing it against the four white-box baselines: SIF, LAEQ, GLiR, and IA. Table~\ref{tab:comparison_mia_metrics} reports AUC, TPR at 1\% and 0.1\% FPR, and pAUC for FPR $\leq$ 1\%. Complementary ROC curves are shown in Figure~\ref{fig:leakboost_vs_baselines_rocs}. 

Across CIFAR-10 and CIFAR-100, LeakBoost exhibits highly variable performance depending on the target architecture. The most striking improvements are observed for ViT-4 and AlexNet. On CIFAR-10, LeakBoost boosts the AUC from $\sim$0.53--0.58 for all baselines to 0.811 on ViT-4 and 0.885 on AlexNet. This gain translates into an \textbf{order-of-magnitude improvement} in low-FPR detection: TPR@1\% increases from $\leq$1.3\% in the baselines to 11.8\% for ViT-4 and 20.15\% for AlexNet, while TPR@0.1\% rises from below 0.2\% to 2--3\%. On CIFAR-100, LeakBoost achieves similar trends, with AUCs of 0.838 (ViT-4) and 0.653 (AlexNet) and markedly higher TPRs at both 1\% and 0.1\% FPR. These results highlight that boosting dynamics uncover strong membership signals in transformer-based and shallow convolutional models that remain hidden to static baselines.

\begin{table}[]
\centering
\small
\renewcommand{\arraystretch}{1.3}
\resizebox{\textwidth}{!}{
\begin{tabular}{lcccccccc}
\toprule
\multirow{2}{*}{\textbf{Method}} &
\multicolumn{4}{c}{\textbf{C-10}} &
\multicolumn{4}{c}{\textbf{C-100}} \\
\cmidrule(lr){2-5} \cmidrule(lr){6-9}
& AUC & TPR@1\% & TPR@0.1\% & pAUC@1\% & AUC & TPR@1\% & TPR@0.1\% & pAUC@1\% \\
\midrule
\midrule
\multicolumn{9}{c}{\textbf{ViT-4}} \\
\midrule
SIF & 0.527 & 0.93\% & 0.10\% & 0.000039 & 0.559 & 1.47\% & 0.12\% & 0.000066 \\
LAEQ & 0.503 & 1.01\% & 0.14\% & 0.000051 & 0.502 & 1.15\% & 0.18\% & 0.000066 \\
GLiR & 0.534 & 0.91\% & 0.06\% & 0.000039 & 0.624 & 1.37\% & 0.14\% & 0.000066 \\
IA & 0.495 & 0.76\% & 0.06\% & 0.000034 & 0.564 & 1.49\% & 0.15\% & 0.000070 \\
\textbf{\attackname{} (Ours.)}& \textbf{0.811} & \textbf{11.80\%} & \textbf{2.39\%} & \textbf{0.000739} & \textbf{0.838} & \textbf{10.60\%} & \textbf{1.42\%} & \textbf{0.000594} \\
\midrule
\multicolumn{9}{c}{\textbf{AlexNet}} \\
\midrule
SIF & 0.568 & 0.93\% & 0.13\% & 0.000046 & 0.670 & 0.98\% & 0.11\% & 0.000051 \\
LAEQ & 0.572 & 0.95\% & 0.06\% & 0.000037 & 0.669 & 1.51\% & 0.12\% & 0.000073 \\
GLiR & 0.578 & 1.28\% & 0.11\% & 0.000060 & \textbf{0.675} & 1.56\% & 0.18\% & 0.000085 \\
IA & 0.501 & 0.78\% & 0.00\% & 0.000039 & 0.577 & 1.36\% & 0.14\% & 0.000069 \\
\textbf{\attackname{} (Ours.)}& \textbf{0.885} & \textbf{20.15\%} & \textbf{3.05\%} & \textbf{0.001099} & 0.653 & \textbf{30.49\%} & \textbf{3.38\%} & \textbf{0.001701} \\
\midrule
\multicolumn{9}{c}{\textbf{DenseNet}} \\
\midrule
SIF & \textbf{0.558} & 1.15\% & 0.08\% & 0.000054 & \textbf{0.642} & 1.01\% & 0.05\% & 0.000056 \\
LAEQ & 0.549 & 1.25\% & 0.13\% & 0.000058 & 0.635 & 1.30\% & 0.17\% & 0.000063 \\
GLiR & 0.557 & 1.24\% & 0.15\% & 0.000061 & 0.637 & 1.53\% & 0.23\% & 0.000076 \\
IA & 0.540 & 1.23\% & 0.11\% & 0.000062 & 0.561 & 1.11\% & 0.13\% & 0.000051 \\
\textbf{\attackname{} (Ours.)}& 0.546 & \textbf{2.24\%} & \textbf{0.37\%} & \textbf{0.000122} & 0.591 & \textbf{1.54\%} & \textbf{0.28\%} & \textbf{0.000080} \\
\midrule
\multicolumn{9}{c}{\textbf{ResNet-18}} \\
\midrule
SIF & \textbf{0.586} & 1.10\% & \textbf{0.13\%} & 0.000055 & \textbf{0.743} & 1.10\% & 0.14\% & 0.000055 \\
LAEQ & 0.548 & 1.17\% & 0.06\% & 0.000047 & 0.677 & 1.40\% & 0.08\% & 0.000063 \\
GLiR & 0.584 & 1.18\% & 0.13\% & 0.000051 & 0.718 & 1.62\% & 0.15\% & 0.000074 \\
IA & 0.560 & \textbf{1.25\%} & 0.11\% & \textbf{0.000062} & 0.620 & 1.66\% & 0.21\% & 0.000084 \\
\textbf{\attackname{} (Ours.)}& 0.530 & 0.95\% & 0.08\% & 0.000047 & 0.678 & \textbf{1.86\%} & \textbf{0.28\%} & \textbf{0.000099} \\
\bottomrule
\end{tabular}
}
\caption{Comparison of membership inference attacks on CIFAR-10 (C-10) and CIFAR-100 (C-100). Metrics: AUC, TPR at 1\% FPR, TPR at 0.1\% FPR, and pAUC for FPR $\leq$ 1\%. Best values in each block are in \textbf{bold}.}
\label{tab:comparison_mia_metrics}
\end{table}

Performance gains are smaller on deeper CNNs. For DenseNet, LeakBoost lags behind in AUC (0.546 on CIFAR-10 vs.\ 0.558 for baselines, and 0.591 on CIFAR-100 vs.\ $\sim$0.64 for baselines), though it achieves a small edge in low-FPR TPRs. ResNet-18 shows a mixed pattern: while LeakBoost reduces the overall AUC on CIFAR-10 (0.530 vs.\ $\sim$0.58), it provides slight gains on CIFAR-100 in the extreme low-FPR regime, with TPR@0.1\% reaching 0.28\% compared to $\leq$0.21\% for baselines. Importantly, while some baselines show marginally higher AUCs in these cases, the differences are relatively minor compared to the dramatic improvements LeakBoost achieves for ViT-4 and AlexNet. Moreover, occasional degradations in AUC are not accompanied by meaningful drops in TPRs at low FPR, which are our primary metrics of interest. This indicates that LeakBoost remains competitive even where it does not clearly outperform, while still excelling in the most privacy-critical operating regime.

\textbf{Takeaways.} 
From the full evaluation, we conclude that: 
(i) LeakBoost delivers dramatic improvements for ViT-4 and AlexNet, raising AUC and especially low-FPR detection rates by an order of magnitude compared to baselines; 
(ii) For deeper CNNs such as DenseNet and ResNet-18, benefits are minor or inconsistent, limited mostly to slight gains in the extreme low-FPR regime.

\subsection{Boosting other detectors and component analysis}
\begin{figure*}[ht]
    \centering
    \includegraphics[width=\textwidth]{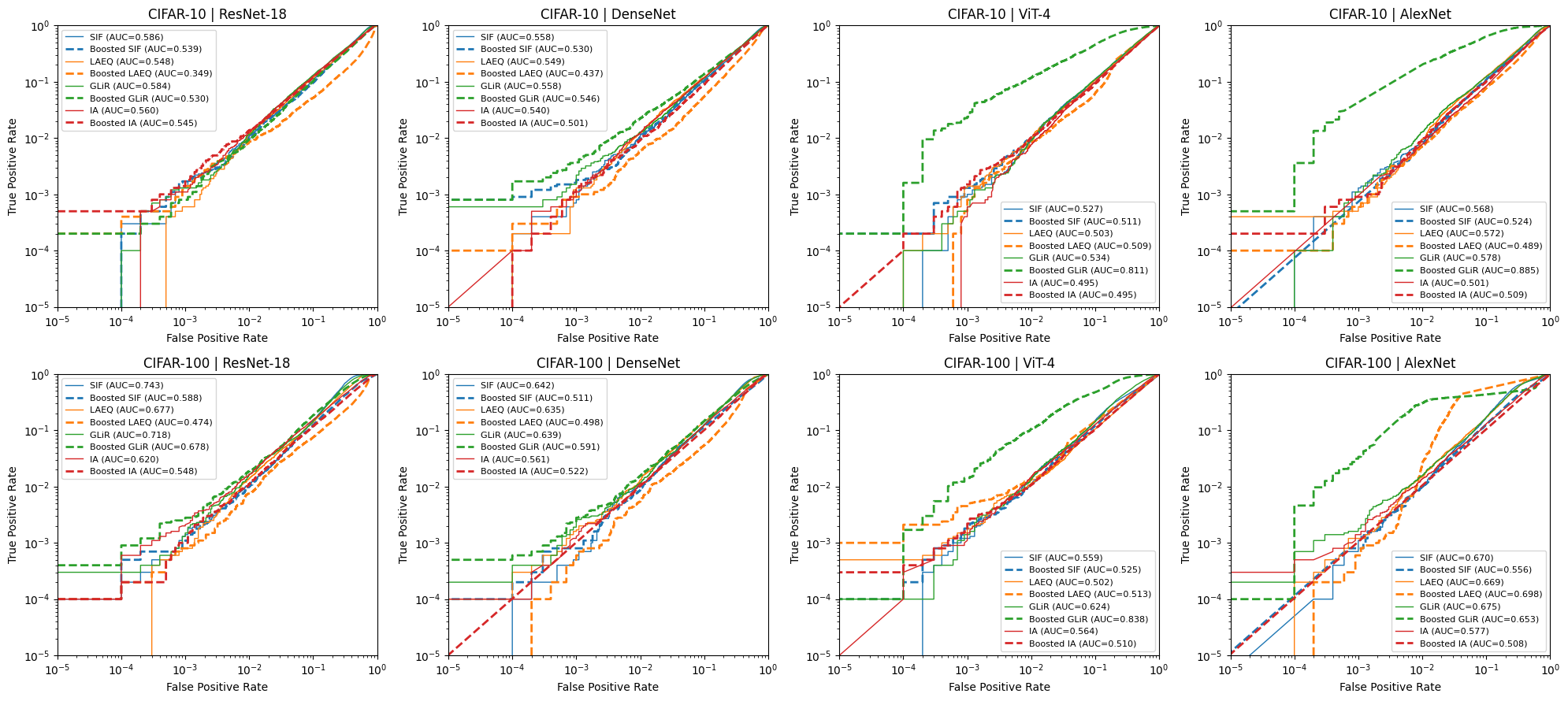}
    \caption{ROC comparison: baseline (solid) vs.\ boosted (dashed) for SIF, LAEQ, IA, and GLiR across CIFAR-10/100 and four architectures. Boosting substantially helps GLiR (notably on ViT-4 and AlexNet) but yields small or negative changes for SIF/LAEQ/IA.}
    \label{fig:e3_roc}
\end{figure*}

\begin{figure*}[ht]
    \centering
    \includegraphics[width=\textwidth]{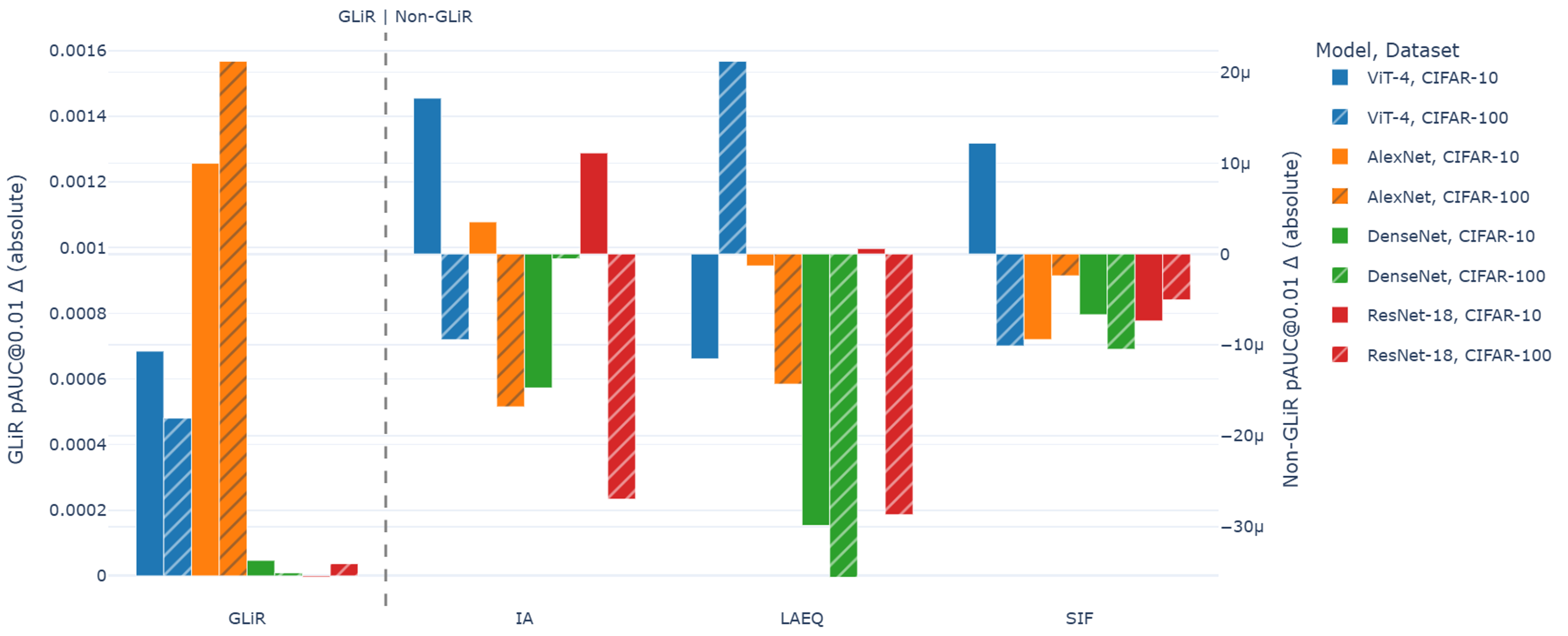}
    \caption{Absolute $\Delta$\,pAUC@1\% from boosting. Bars to the left (``GLiR'') show consistent positive improvements ($\sim 10^{-3}$), while non-GLiR detectors display near-zero or negative shifts (mostly within $\pm 10^{-5}$ to $10^{-4}$).}
    \label{fig:e3_pauc}
\end{figure*}

We apply the LeakBoost interrogation step to alternative detectors (SIF, LAEQ, and IA) to test whether boosting generally helps white-box MIAs, or whether the gains in the previous experiment are specific to GLiR. Figure~\ref{fig:e3_roc} compares ROC curves for each architecture with and without boosting, and Figure~\ref{fig:e3_pauc} summarizes the absolute change in pAUC for FPR~$\leq$~1\%.

\textbf{Boosting strongly favors GLiR.}
Across models and datasets, GLiR consistently benefits from boosted inputs, with $\Delta$\,pAUC@1\% typically on the order of $10^{-3}$ (most pronounced on ViT-4 and AlexNet). In contrast, boosting SIF/LAEQ/IA yields only marginal and often negative changes, usually within $\pm 10^{-5}$ to $10^{-4}$, with no systematic improvement. This pattern holds in the low-FPR tails: LeakBoosted GLiR raises TPR@1\% and TPR@0.1\% by multiples, whereas boosted SIF/LAEQ/IA rarely shift the tail meaningfully.

\textbf{Why GLiR gains, others do not.}
GLiR’s statistic depends explicitly on \emph{gradients}, so driving a short optimization (our interrogation) directly amplifies gradient-based separability between members and non-members. In contrast, SIF, LAEQ, and IA rely primarily on static features that are less aligned with the transient optimization dynamics. As a result, perturbations that meaningfully reshape the local gradient field (helping GLiR) do not consistently increase class separation in those static signals.

\textbf{Takeaways.}
(i) Boosting is not a universal plug-in: its large wins are \emph{detector-dependent} and materialize when the detector reads gradient information. 
(ii) Where non-gradient detectors do improve, the gains are small and inconsistent, suggesting limited utility of interrogation for purely activation/loss-based signals.

\subsection{Hyperparameters Sensitivity Analysis}

To assess the robustness of our interrogation procedure, we conduct a comprehensive sensitivity analysis over its main hyperparameters: the number of optimization steps, the learning rate, and the use of value clipping. In addition, we examine how different layer groups (early, mid, late, and all layers combined) contribute to the strength of the extracted signals. For this analysis, we focus on two representative architectures, ViT-4 and AlexNet, which in our main experiments were found to be the most affected by the attack. This choice enables us to highlight how hyperparameters interact not only with the attack dynamics but also with architectural differences, contrasting transformer-based representations against classical convolutional networks. The chosen hyperparameters appear in Tables \ref{tab:cifar10_leakboost_config} and \ref{tab:cifar100_leakboost_config}.

\begin{figure}[H]
    \centering
    \makebox[\textwidth][c]{%
        \includegraphics[width=1\textwidth]{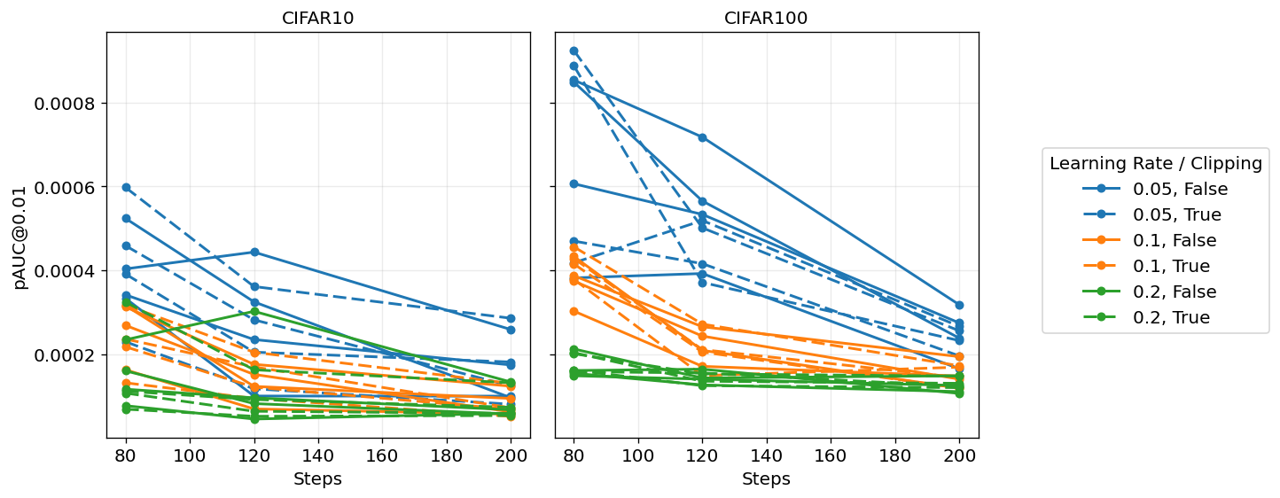}%
    }
    \caption{Sensitivity of pAUC@0.01 to learning rate, clipping, and steps for ViT-4 on CIFAR-10 and CIFAR-100. Each line corresponds to one hyperparameter configuration, showing how performance evolves with increasing optimization steps.}
    \label{fig:vit_line}
\end{figure}

\begin{figure}[ht]
    \centering
    \makebox[\textwidth][c]{%
        \begin{minipage}{1\textwidth}
            \centering
            \begin{subfigure}{1\linewidth}
                \includegraphics[width=\linewidth]{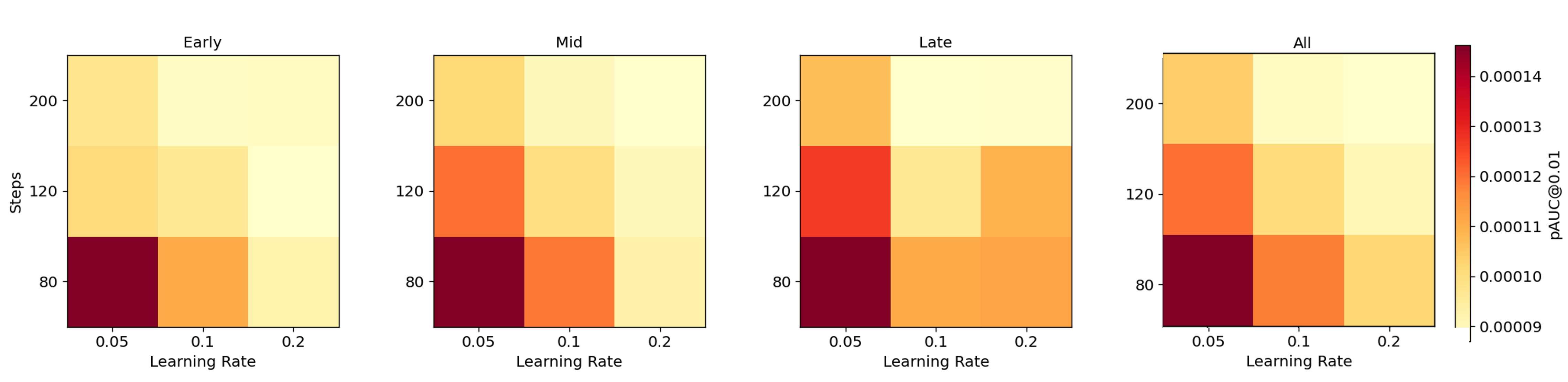}
                \caption{}
                \label{fig:plotA}
            \end{subfigure}
            \vfill
            \begin{subfigure}{1\linewidth}
                \includegraphics[width=\linewidth]{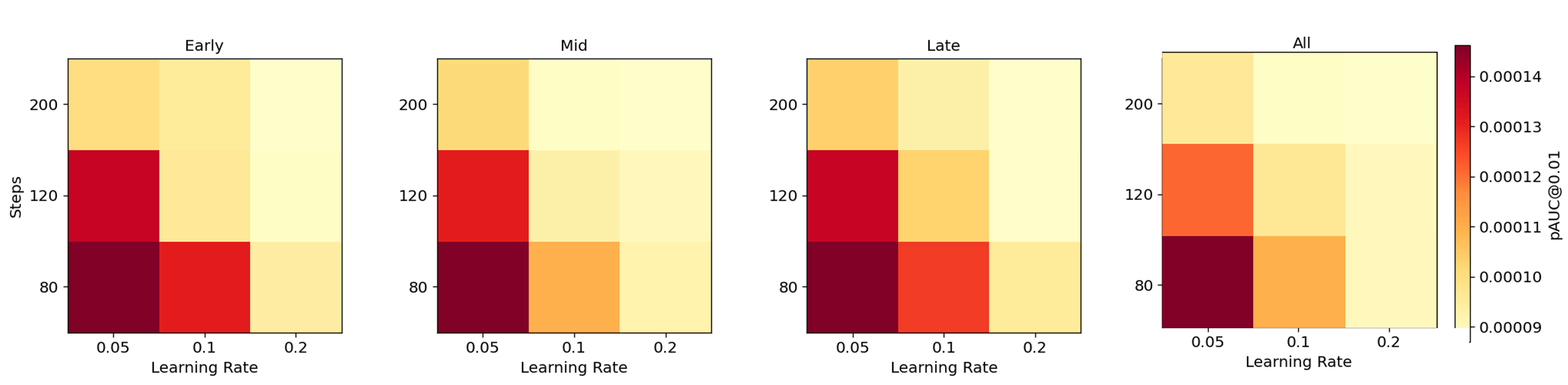}
                \caption{}
                \label{fig:plotB}
            \end{subfigure}
        \end{minipage}%
    }
    \caption{Heatmaps of pAUC@0.01 for ViT-4 on CIFAR-10 (a) and CIFAR-100 (b) across learning rates and optimization steps, shown separately for early, mid, late, and all layers.}
    \label{fig:vit_heatmaps}
\end{figure}

\begin{figure*}[th]
    \centering
    \makebox[\textwidth][c]{%
        \includegraphics[width=1\textwidth]{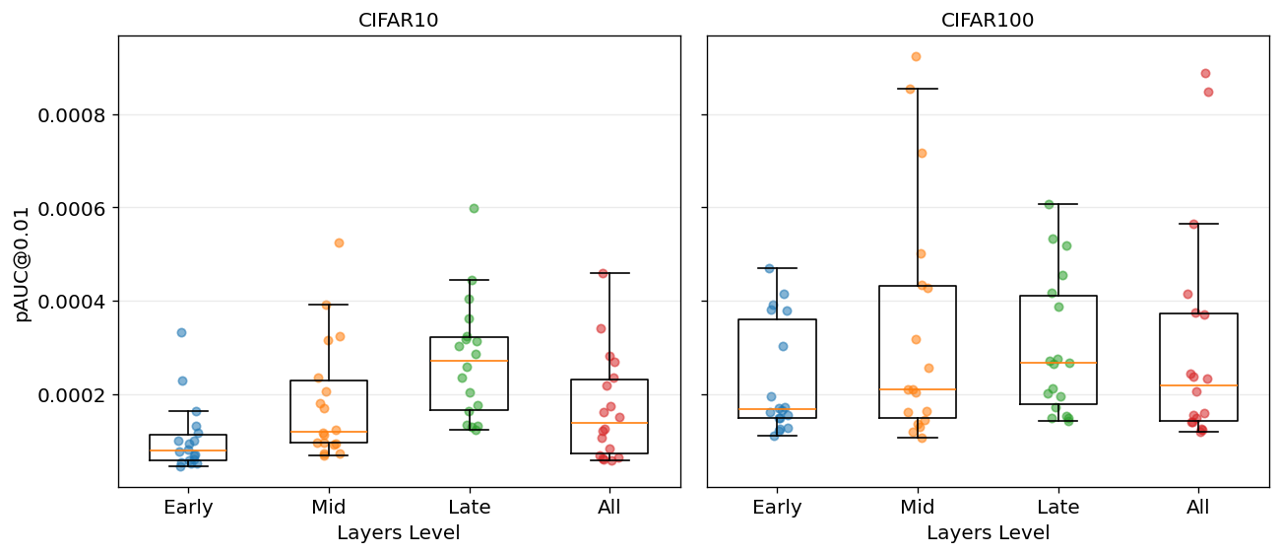}%
    }
    \caption{Distribution of pAUC@0.01 values for ViT-4 across early, mid, late, and all layers on CIFAR-10 and CIFAR-100. The results show that leakage is weak in early layers but becomes substantially stronger and more variable in deeper representations.}
    \label{fig:vit_layers}
\end{figure*}

\textbf{ViT-4.  }
The ViT-4 results (Figures~\ref{fig:vit_line}, \ref{fig:vit_heatmaps}, and \ref{fig:vit_layers}) reveal a highly consistent pattern across both CIFAR-10 and CIFAR-100. The number of optimization steps has a clear monotonic effect: membership advantage is strongest at short optimization (80 steps) and steadily declines with longer runs. By 200 steps, the signal is almost entirely diminished, suggesting that excessive optimization smooths out the subtle discrepancies between members and non-members that our method is designed to exploit. This pattern holds across both datasets and across most learning-rate settings.

The learning rate itself also plays an important role. Smaller values (0.05) consistently yield the highest partial AUC scores, particularly at shorter step counts, while larger values (0.1--0.2) accelerate the degradation of performance. Thus, there appears to be a regime of ``gentle'' optimization---short runs with low learning rates---in which the attack is most effective. Interestingly, the application of clipping does not produce consistent gains, with differences between clipped and unclipped runs being marginal at best. This suggests that clipping is not a decisive factor in the ViT setting.

Layer-level analysis provides further insights. As shown in Figure~\ref{fig:vit_layers}, the results reveal a clear dependency of membership leakage on layer depth. On both CIFAR-10 and CIFAR-100, early layers yield weak signals, with consistently low pAUC values and limited variance. In contrast, mid and late layers exhibit substantially stronger leakage. For CIFAR-10, late layers achieve the highest median performance, while mid layers provide competitive results but with more variability. On CIFAR-100, this trend is amplified: mid layers show a very wide spread, occasionally reaching the strongest pAUC values, whereas late layers produce more stable and consistently high results. Interestingly, aggregating all layers does not outperform the best subsets, suggesting that signals from weak layers can dilute the strength of informative ones. Overall, these findings highlight that the most discriminative membership signals in ViT models emerge from deeper representations, but that intermediate layers can sometimes expose even stronger individual leakage patterns.

\textbf{AlexNet. } 
We next examine AlexNet, which demonstrates a markedly different sensitivity pattern compared to ViT. 

\begin{figure}[H]
    \centering
    \makebox[\textwidth][c]{%
        \includegraphics[width=1\textwidth]{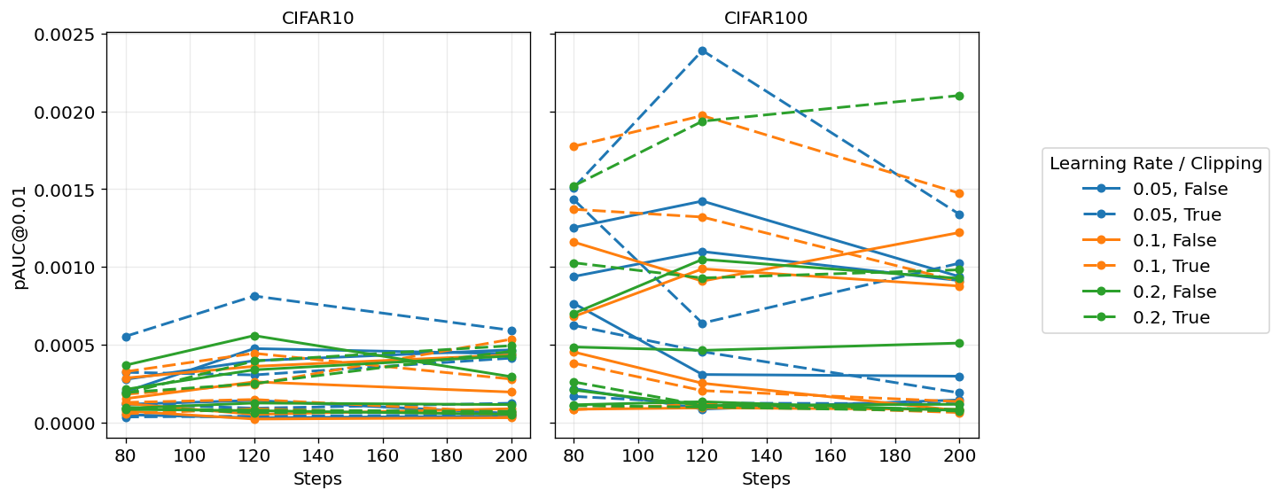}%
    }
    \caption{Sensitivity of pAUC@0.01 to learning rate, clipping, and steps for AlexNet on CIFAR-10 and CIFAR-100. Each line corresponds to one hyperparameter configuration, showing how performance evolves with increasing optimization steps.}
    \label{fig:alexnet_line}
\end{figure}

The line plots in Figure~\ref{fig:alexnet_line} show that, unlike ViT, where performance consistently decreases with more optimization steps, AlexNet exhibits highly unstable behavior. On CIFAR-10, the attack achieves relatively uniform performance across different learning rates and clipping settings, with only mild fluctuations as the number of steps increases. However, on CIFAR-100, the attack becomes far less predictable: some configurations achieve strong leakage at intermediate steps (most notably at $120$ steps with learning rate $0.05$ and clipping enabled), while others degrade substantially, resulting in a tangled set of trajectories without a clear monotonic trend. This suggests that AlexNet’s feature geometry makes the interrogation procedure more sensitive to hyperparameter tuning, particularly on more complex datasets.

\begin{figure}[ht]
    \centering
    \makebox[\textwidth][c]{%
        \begin{minipage}{1\textwidth}
            \centering
            \begin{subfigure}{1\linewidth}
                \includegraphics[width=\linewidth]{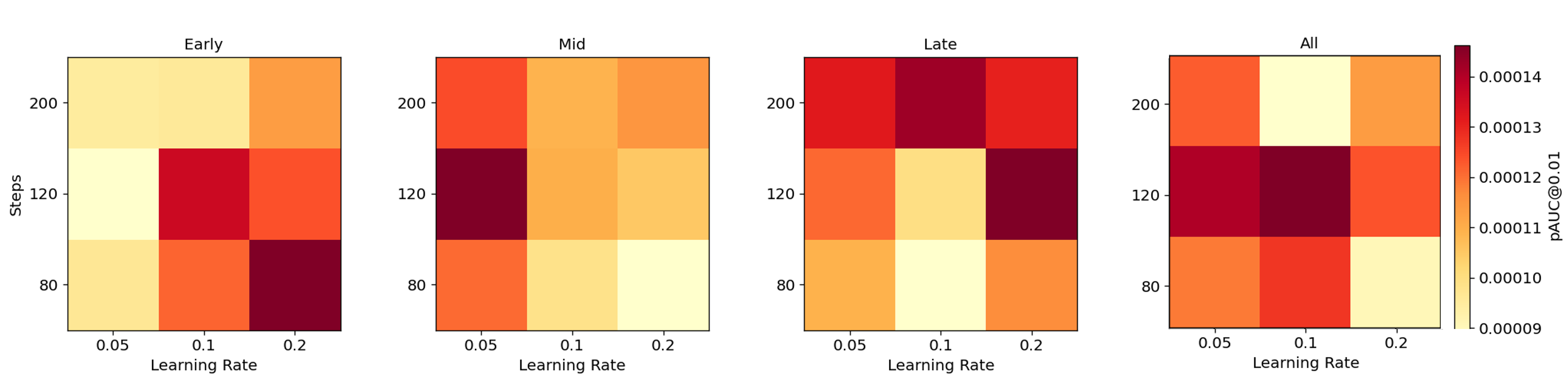}
                \caption{}
                \label{fig:plotA}
            \end{subfigure}
            \vfill
            \begin{subfigure}{1\linewidth}
                \includegraphics[width=\linewidth]{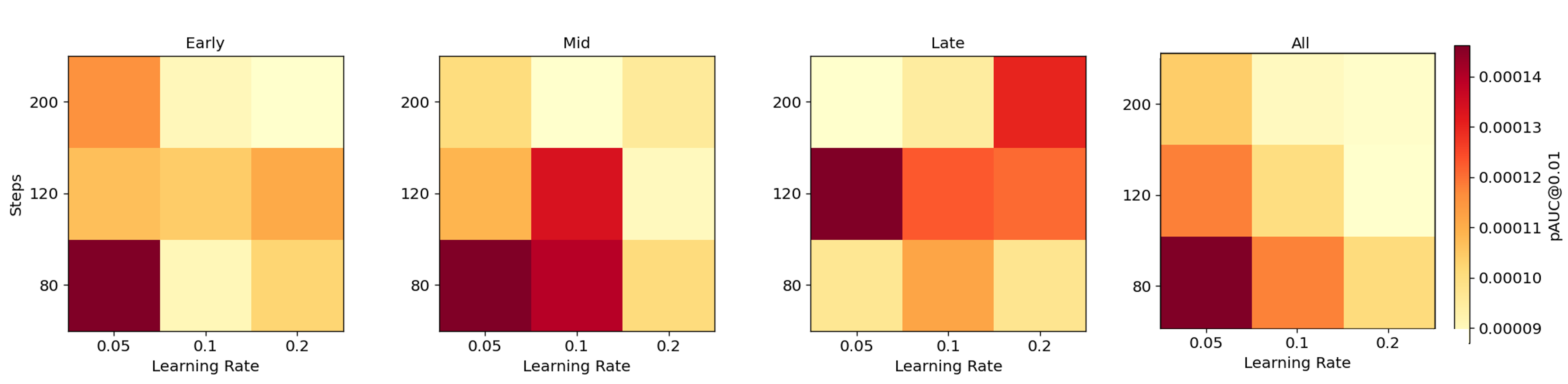}
                \caption{}
                \label{fig:plotB}
            \end{subfigure}
        \end{minipage}%
    }
    \caption{Heatmaps of pAUC@0.01 for AlexNet on CIFAR-10 (a) and CIFAR-100 (b) across learning rates and optimization steps, shown separately for early, mid, late, and all layers.}
    \label{fig:alexnet_heatmaps}
\end{figure}

The heatmaps in Figure~\ref{fig:alexnet_heatmaps} reinforce this observation. On CIFAR-10, leakage is generally weak and relatively flat across learning rates and steps, with only isolated “hot spots” appearing in late and mid layers. In contrast, CIFAR-100 reveals stronger and more variable leakage, especially in late layers where low learning rates combined with small step counts yield the highest pAUC values. This inconsistency across datasets indicates that AlexNet’s vulnerability to interrogation is not robust but depends on both the dataset and the exact hyperparameter regime.

\begin{figure*}[th]
    \centering
    \makebox[\textwidth][c]{%
        \includegraphics[width=1\textwidth]{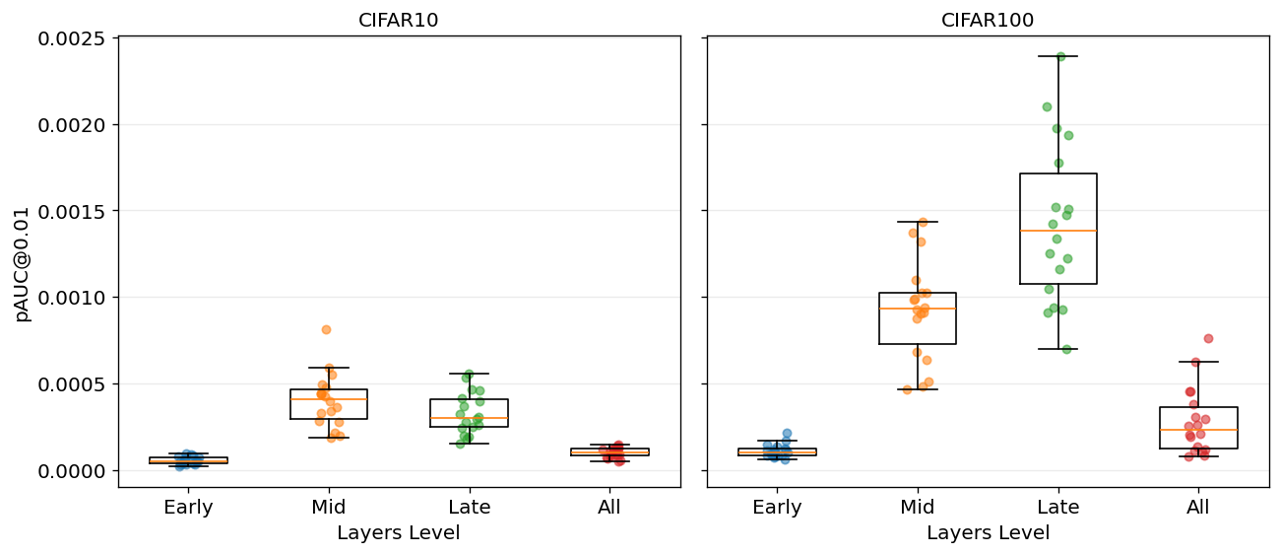}%
    }
    \caption{Distribution of pAUC@0.01 values for AlexNet across early, mid, late, and all layers on CIFAR-10 and CIFAR-100. The results show that leakage is weak in early layers but becomes substantially stronger and more variable in deeper representations.}
    \label{fig:alexnet_layers}
\end{figure*}

Finally, the layer-level analysis in Figure~\ref{fig:alexnet_layers} highlights the key role of depth. On CIFAR-10, early and all-layer settings yield near-random leakage, while mid and late layers provide the strongest signals, though still weaker than in ViT. On CIFAR-100, the trend is much more pronounced: late layers leak substantially more membership information, with very high variance across runs, while early layers remain almost entirely uninformative. This aligns with the intuition that deeper convolutional layers in AlexNet encode more memorization of the training set, but also reveals that combining all layers tends to dilute these signals. Overall, AlexNet shows that the effectiveness of our interrogation method is highly architecture-dependent. While ViT yields stable improvements with deeper layers, AlexNet’s leakage emerges sporadically and only under specific hyperparameter configurations.

\textbf{Takeaways. }
From the sensitivity study, we conclude that: 
(i) membership leakage depends strongly on both model architecture and hyperparameter choice, with ViT showing relatively stable patterns while AlexNet remains unstable and dataset-dependent; 
(ii) layer depth plays a decisive role, as early layers are weak, mid and late layers provide stronger signals, and combining all layers may dilute performance; 
(iii) The most efficient configuration consistently involves a learning rate of $0.05$ with clipping enabled, underscoring the importance of careful tuning to realize the attack’s full potential.

\subsection{Overall Discussion}

Across all evaluations, LeakBoost transforms passive white-box attacks into an active probing framework. 
Its benefits are architectural: transformers and shallow CNNs leak more when interrogated, 
whereas deeper, highly regularized models leak less. 
Across datasets, the gains are also consistent: on \textbf{CIFAR-100}, LeakBoost improves performance in \emph{all} evaluated settings, and on \textbf{CIFAR-10} it improves performance in \emph{most} cases. 
Because LeakBoost is detector-agnostic and computationally lightweight, it can serve as both an offensive tool for analyzing worst-case privacy leakage 
and a defensive diagnostic for quantifying model memorization under full access.

LeakBoost thus advances the study of membership inference toward a dynamic regime—one where privacy risk is revealed not only by static statistics but by how a model \emph{responds} when questioned about its own knowledge.

\subsection{\textbf{Defenses and Mitigations.}}
Defenses against membership inference generally fall into \emph{training-time} mechanisms that reduce memorization (e.g., strong regularization and augmentation \cite{MIAShokri2017, kaya2021does}, early stopping \cite{song2021systematic}, and differentially private training \cite{abadi2016deep}) and \emph{inference-time} mitigations that restrict or obfuscate the signals available to an attacker \cite{jia2019memguard}. While differential privacy provides principled guarantees, it often introduces non-trivial accuracy and tuning trade-offs. We therefore focus on \emph{inference-time} mitigations that better match deployment constraints (e.g., protecting already-trained models) and align with the threat model by controlling what information is exposed to an attacker. We evaluate an activation-obfuscation variant in which the adversary observes noisy intermediate representations (layer normalization followed by additive Gaussian noise). Empirically, this perturbation yields negligible degradation for LeakBoost when paired with GLiR, indicating that LeakBoost+GLiR can effectively \emph{bypass} activation-level obfuscation at the tested noise levels.

\section{Conclusion}

This work introduced \textbf{LeakBoost}, a modular framework for amplifying membership signals through \emph{activation-space interrogation}. 
Unlike prior white-box attacks that rely on static gradients or losses, LeakBoost actively probes a target model by optimizing a perceptual loss that aligns internal representations. 
This process generates an \emph{interrogation image} that makes latent memorization effects more explicit, enabling existing detectors—such as GLiR—to operate with significantly higher accuracy, especially in the low–false-positive regime.

Extensive experiments across architectures (ResNet-18, DenseNet, AlexNet, and ViT-4) and datasets (CIFAR-10 and CIFAR-100) show that LeakBoost substantially improves attack success while remaining computationally lightweight and detector-agnostic. 
Ablation and sensitivity analyses further reveal that membership leakage depends strongly on model architecture and interrogation design, highlighting the link between representational geometry and privacy risk.

By reframing membership inference as an active interrogation process, LeakBoost provides a new lens for studying how neural networks memorize and reveal their training data. 
Beyond its use as an attack, it serves as a practical \emph{privacy assessment tool} for evaluating model vulnerability under white-box access. 
Future work will explore adapting this framework to other modalities (text, audio) and to defense evaluation under differential privacy and adversarial training regimes.

\bibliographystyle{ACM-Reference-Format}
\bibliography{sample-base}


\appendix
\section{Appendix}

\subsection{Target Models}
This section reports the training, validation, and test accuracies of the target models used in our experiments on CIFAR-10 and CIFAR-100.

\begin{table}[h]
\centering
\small
\setlength{\tabcolsep}{6pt}
\begin{tabular}{lcccccc}
\toprule
\multirow{2}{*}{\textbf{Model}} & 
\multicolumn{3}{c}{\textbf{CIFAR-10}} & 
\multicolumn{3}{c}{\textbf{CIFAR-100}} \\
\cmidrule(lr){2-4} \cmidrule(lr){5-7}
& Train & Test & Val & Train & Test & Val \\
\midrule
ResNet-18 & 0.9984 & 0.8947 & 0.9068 & 0.9694 & 0.5987 & 0.6148 \\
AlexNet   & 0.9402 & 0.8005 & 0.8064 & 0.7388 & 0.3972 & 0.3836 \\
DenseNet  & 0.9850 & 0.8960 & 0.9008 & 0.8838 & 0.6088 & 0.5936 \\
ViT-4     & 0.7212 & 0.6757 & 0.6664 & 0.5978 & 0.3775 & 0.3640 \\
\bottomrule
\end{tabular}
\caption{Accuracy of the target models on CIFAR-10 and CIFAR-100 datasets.}
\label{tab:target_models_acc}
\end{table}

\subsection{Hyperparameters}
This section summarizes the layer groupings and optimization hyperparameters used for LeakBoost across architectures and datasets.

\begin{table}[H]
\centering
\small
\begin{tabular}{lcccc}
\hline
Model & Layer level & Steps & LR & Clipping \\
\hline
ResNet18 & All   & 80  & 0.05 & True \\
DenseNet & Late  & 80  & 0.05 & True \\
ViT4     & Late   & 80  & 0.05 & True \\
AlexNet  & Mid   & 120 & 0.05 & True \\
\hline
\end{tabular}
\caption{Best configurations for LeakBoost attacks on CIFAR-10 (validation set).}
\label{tab:cifar10_leakboost_config}
\end{table}

\begin{table}[H]
\centering
\small
\begin{tabular}{lcccc}
\hline
Model & Layer level & Steps & LR & Clipping \\
\hline
ResNet18 & Late  & 120 & 0.10 & True \\
DenseNet & Late  & 120 & 0.05 & True \\
ViT4     & Mid   & 80  & 0.05 & True \\
AlexNet  & Late  & 120 & 0.05 & True \\
\hline
\end{tabular}
\caption{Best configurations for LeakBoost attacks on CIFAR-100 (validation set).}
\label{tab:cifar100_leakboost_config}
\end{table}

\begin{table}[t]
\centering
\small
\begin{tabular}{lp{2.6cm}p{2.6cm}p{2.6cm}p{3.0cm}}
\hline
Model & Early layers & Mid layers & Late layers & All layers \\
\hline
ResNet18 &
Input conv, block group 1 &
Block groups 2--3 &
Block group 4, global avg pool &
All residual blocks and pooling \\
DenseNet &
Initial conv, dense block 1 &
Dense block 2 &
Dense block 3, classifier &
All dense blocks and classifier \\
ViT4 &
Patch embedding, encoder blocks 0--1 &
Encoder blocks 2--4 &
Encoder block 5, layer norm, head &
All encoder blocks and output layers \\
AlexNet &
Conv layers 1--2 &
Conv layers 3--4 &
Conv layer 5, classifier &
All conv and classifier layers \\
\hline
\end{tabular}
\caption{Layer groups are defined at the architectural level for clarity; exact mappings to implementation-specific module names are provided in the code.}
\label{tab:layers_subsets}
\end{table}

\end{document}